\documentclass{article}

\usepackage{amsmath}
\usepackage{microtype}
\usepackage{graphicx}
\usepackage{subfigure}
\usepackage{amsfonts}
\usepackage{booktabs} 

\usepackage{hyperref}

\usepackage[accepted]{icml2019}

\icmltitlerunning{Neural Heterogeneous Scheduler}

\begin{document}

\twocolumn[
\icmltitle{Neural Heterogeneous Scheduler}



\icmlsetsymbol{equal}{*}

\begin{icmlauthorlist}
\icmlauthor{Tegg Taekyong Sung}{to,equal}
\icmlauthor{Valliappa Chockalingam}{to,alb,equal}
\icmlauthor{Alex Yahja}{to}
\icmlauthor{Bo Ryu}{to}
\end{icmlauthorlist}

\icmlaffiliation{to}{EpiSys Science, San Diego, USA}
\icmlaffiliation{alb}{Department of Computer Science, University of Alberta, Edmonton, Canada}

\icmlcorrespondingauthor{Tegg Taekyong Sung}{tegg@episyscience.com}
\icmlcorrespondingauthor{Valliappa Chockalingam}{valli@episyscience.com}

\icmlkeywords{Machine Learning, ICML}

\vskip 0.3in
]



\printAffiliationsAndNotice{\icmlEqualContribution} 

\begin{abstract}
Access to parallel and distributed computation has enabled researchers and developers to improve  algorithms and performance in many applications. Recent research has focused on next generation special purpose systems with multiple kinds of coprocessors, known as heterogeneous system-on-chips (SoC)~\cite{uhrie2019machine}. In this paper, we introduce a method to intelligently schedule--and learn to schedule--a stream of tasks to available processing elements in such a system. We use deep reinforcement learning enabling complex sequential decision making and empirically show that our reinforcement learning system provides for a viable, better alternative to conventional scheduling heuristics with respect to minimizing execution time.
\end{abstract}

\section{Introduction}
\label{sec:introduction}
The deep learning renaissance was made possible in part due to the superbly curated ImageNet~\cite{deng2009imagenet} and the groundbreaking image classifiers in the 2012 ImageNet Challenge~\cite{krizhevsky2012imagenet}. However, this was also owing to the increased availability and affordability of massively parallel graphics processing units (GPUs) on which core operations of neural networks, such as matrix multiplication, could be very efficient. Recent research in heterogeneous, reconfigurable hardware points to a sea-change in hardware on which further deep learning innovations can be enabled. In preparation for these innovations, it is imperative to examine distinct performance characteristics afforded by new hardware, for instance, the execution time, power and energy consumption of custom ASICs, or FPGAs. To maximize performance of different applications, scheduling tasks to hardware processing elements suited for running them and finding optimal combinations is necessary~\cite{gupta2017dypo}. Consequently, how to optimally schedule operations becomes the main research topic which is generally known as an NP-hard problem. 

Traditionally, heuristics or approximate algorithms can make the scheduling problem more tractable. Heuristic schedulers however generally lack learning components. They actively search optimal actions given the fixed profiles of tasks and processing elements, however their performance likely deteriorates when unexpected information is encountered during scheduling. To avoid such undesired results and go beyond heuristics, in this paper, we make use of deep reinforcement learning (DRL) which provides a powerful and adaptive way to solve complex sequential decision making problems. We argue that optimal distribution of tasks into available processing elements can be achieved using learning algorithms. Here, we consider heterogeneous processing elements (a mixture of CPUs, GPUs, ASICs, and other cores) and tasks that have dependencies: the scheduling agent must learn how to schedule the tasks given that some tasks may require other tasks to have already run. This makes the problem challenging due to long-term dependencies and partial observability. Moreover, without pre-emption, the dependencies entail that the agent cannot choose scheduling actions at every time step: it chooses an action only when assigned tasks are completed and new tasks ready to be scheduled appear. The following sections describe the details of the simulation environment which has input parameters of a job consisting of a set of tasks and of a resource matrix specifying the performance and communication specifications of the processing elements. Next, we formalize the reinforcement learning (RL) setting and describe the policy-based algorithm we use to tackle the sequential decision-making problem. In the experiments, we compare our model, which we interchangeably refer to as Deep Resource Manager (DRM) or Neural Heterogeneous Scheduler, with baselines to show that the performance of the Deep RL agent we introduce is better than those of different heuristic scheduling algorithms. We also provide saliency maps and GANTT chart visualizations during the learning process for analyses of the agent's decisions.

\section{Background}
\label{sec:background}
Deep reinforcement learning (Deep RL) has been successfully applied to several domains such as robotics~\cite{levine2016end} and games~\cite{silver2017mastering,vinyals2019alphastar}. Most successful RL applications stand in the usual RL framework of Markov decision processes. However, in our case, actions can take various lengths of time to complete. Scheduling chip processors in real-world applications involves a continually filling stream of tasks where many activities progress simultaneously. Action decisions are only performed when tasks are ready to be scheduled. Given these properties and limitations of the environment, this process can be defined as semi-Markov decision process (SMDP) with temporally-extended actions or options. When an assigned action is not completed, then the agent essentially takes the `no-operation' action. 

Mathematically, the MDP setting can be formalized as a 5-tuple $\langle \mathcal{S}, \mathcal{A}, R, P, \gamma \rangle$~\cite{sutton2018reinforcement,puterman2014markov}. Here, $S$ denotes the state space, $A$, the action space, $R$, the reward signal which is generally defined over states or state-action pairs, $P$, a stochastic matrix specifying transition probabilities to next states given a state and action, and $\gamma \in [0, 1]$, the discount factor. 

Normally, the SMDP framework would involve an option framework, augmenting the action space, but instead, we use simple options here that take no-op actions and hence leave the option framework with preemption of running tasks as future work. 

In addition, the heterogeneous resource management environment is essentially partially observable, because the agent can only observe the tasks ready-to-be-assigned to a processing element. To address this, we augment the state with the other task lists as well (not just the ready list containing the ready-to-be-assigned tasks) and transition to fully-observable problem. Here fully-observable is loosely defined as the environment where all task statuses are represented into a state in every timestep for our learning algorithms.

\section{Proposed Approach}
\label{sec:proposed}

\subsection{Environment Setting}
\label{sec:proposed:environment}
The heterogeneous SoC chip we consider is to be used in various applications, such as WiFi RX/TX or pulse doppler. They are simulated using a discrete-event Domain-Specific System-on-Chip Simulation (DS3) framework developed recently~\cite{arda2019ds3}. As described in~\ref{sec:proposed:environment:simulation}, it is developed with the SimPy library~\cite{Simpy2018} to implement running tasks in a continuous time frame setting. The specifications of the set of tasks and processing elements are written within \texttt{job} and \texttt{resource\_matrix}  files that are described in~\ref{sec:proposed:environment:taskresource}.

\subsubsection{Simulation}
\label{sec:proposed:environment:simulation}
We consider a simulated environment to start with as in many RL applications. The goal of the agent is to achieve a low time to completion given a set of tasks.

\begin{figure}[ht]
\begin{center}
\centerline{\includegraphics[width=0.75\linewidth]{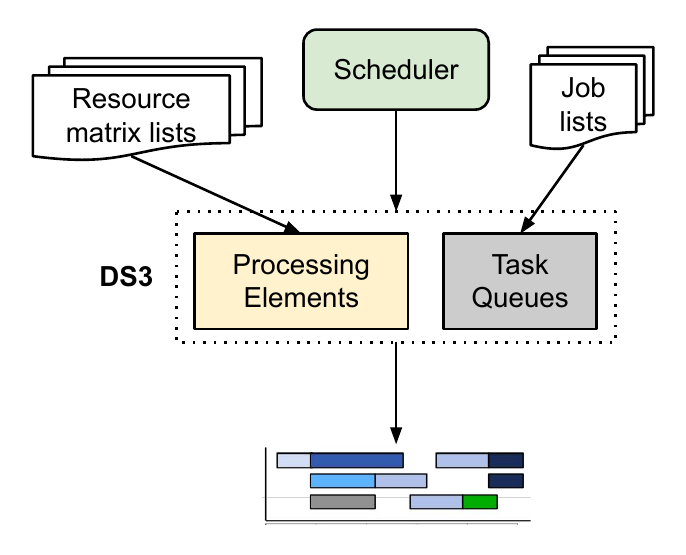}}
\caption{An overall diagram of the DS3 environment which runs with the SimPy discrete-event library. The simulator constructs task queues and processing elements based on the descriptions in the \texttt{job} and \texttt{resource\_matrix} file lists. The scheduler assigns tasks in the ready list to cores in SoC chip.}
\label{fig:ds3}
\end{center}
\vskip -0.2in
\end{figure}

Recently, RL algorithms are usually developed using the OpenAI Gym environment interface~\cite{brockman2016openai} and many assume the Markov decision process framework, whereas we use SimPy environment, which simulates sequential discrete-events for each processing element. Each event represents a task. Each task, upon execution, runs till completion and the scheduler can only choose to schedule tasks in the ready list to processing elements. The simulator is visualized in Figure~\ref{fig:ds3}. 

Prior to running a simulation, the information of processing elements (PEs) and tasks are parsed with \texttt{resource\_matrix} and \texttt{job} text files described in~\ref{sec:proposed:environment:taskresource}. Each PE represents chipsets such as RAM, CPU, GPU, or memory accelerators in heterogeneous SoC and have different execution time, energy and power consumption. In this paper, we only consider the execution time for the performance. 

At the start of a job, all the tasks are fed into the \textit{outstanding list} and the ones that do not have task dependencies are pushed into the \textit{ready list}. Then, the agent assigns a processing element, through an ID, \texttt{PE\_ID}, to tasks that are in the \textit{ready list} and these tasks then proceed to the \textit{running list} when the begin execution. During this time period, the agent chooses a `no-operation' action for the running tasks. Once a task completes, it is moved to the \textit{completed list}. If all the tasks are in the \textit{completed list}, the scheduling episode is finished and the next \texttt{resource\_matrix} and \texttt{job} files in the list are used for the next episode.

\subsubsection{Task and Resource matrix}
\label{sec:proposed:environment:taskresource}
\begin{figure}[ht]
\begin{center}
\centerline{\includegraphics[width=0.5\linewidth]{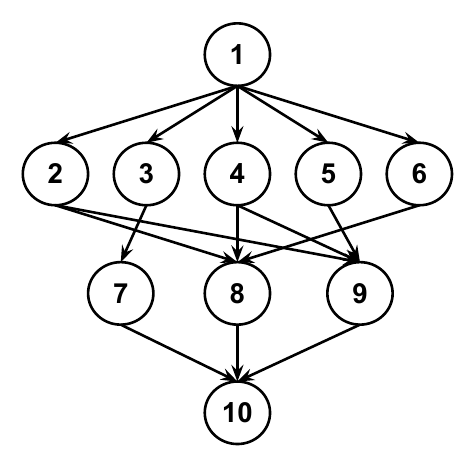}}
\caption{Task dependency visualization of the \texttt{job\_Top} file. Each circle represents task numbers and arrows show task dependencies.}
\label{fig:task-depend}
\end{center}
\vskip -0.2in
\end{figure}

Tasks are constantly generated and a scheduler distributes them to different chipsets in the SoC. We assume tasks have dependencies such as those shown in Figure~\ref{fig:task-depend}. We describe the list of tasks in a \texttt{job} file and the associated processing elements in a \texttt{resource\_matrix} file. Their structures are described in below.

\textbf{Job list}
\label{list:job}
\begin{itemize}
\item \texttt{job\_name} $<$job name$>$
\item \texttt{add\_new\_tasks} $<$number of tasks$>$
\item $<$task name$>$ $<$task ID$>$ $<$task predecessors$>$
\item $<$task name$>$ $<$earliest start time$>$ $<$deadline$>$
\end{itemize}

\textbf{Resource matrix list}
\label{list:resource}
\begin{itemize}
\item \texttt{add\_new\_resource} $<$resource ID$>$ $<$number of tasks$>$
\item $<$task name$>$ $<$performance$>$
\end{itemize}

In a job file, tasks have \texttt{HEAD} and \texttt{TAIL} flags that indicate the start and end. In this paper, we consider 10 tasks of one job with 3 processing elements. However, we add randomization to the resource matrix to train our agent be more robust. This results in differing performances. Performance here refers to the fact that the execution time taken to process a given task in a given processing element varies. The earliest start time, deadline, and performance are all given in units of milliseconds.

\subsection{Algorithm}
\label{sec:algorithm}
In this paper, we develop a new agent using deep reinforcement learning to allocate resources to a heterogeneous SoC, a long-term credit assignment task. As described in the Section~\ref{sec:background}, the environment in its most general form can be thought of as a partially observable SMDP. 

\begin{figure}[ht]
\begin{center}
\centerline{\includegraphics[width=0.5\linewidth]{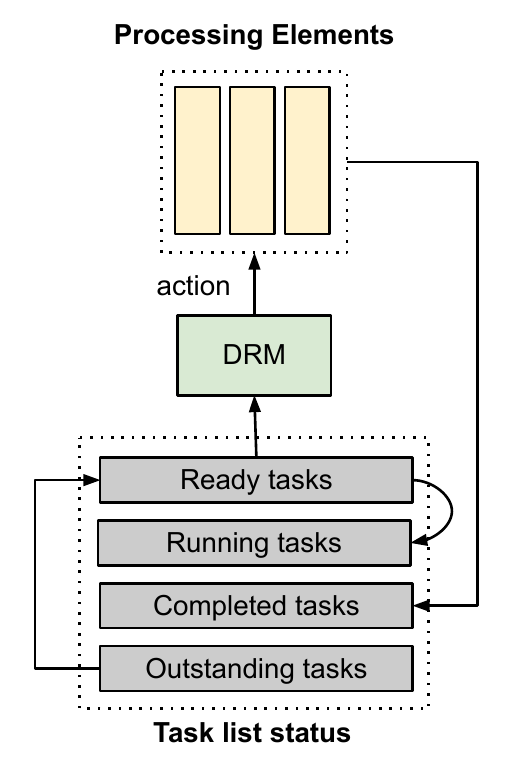}}
\caption{A diagram of the interaction of DRM architecture and DS3 environment. Tasks in the lists are controlled by DS3 environment and DRM assigns every \texttt{PE\_ID} to ready tasks.}
\label{fig:arch}
\end{center}
\vskip -0.2in
\end{figure}

Figure~\ref{fig:arch} shows the interaction between DS3 and DRM scheduler. Task list transitions are controlled by DS3 environment. The scheduler agent takes the tasks from the ready list as an input and assigns each task a \texttt{PE\_ID}. Particularly, our DRM scheduler receives ready tasks but also generates state representations with all the task lists. We convert the task lists and resource matrix to binary vector representations when representing integer values, multi-binary representations for state features that can take on multiple values and concatenate the representation to form the final state representation vector. This information from all the state lists not only addresses partial observability in the DS3 environment, but also gives additional information about the relations between tasks through the task list transitions. 

We use an actor-critic algorithm described in Equation ~\ref{eq:pg}~\cite{sutton2000policy}. The agents action is taken for tasks in ready list. We use a simple reward of -1 per timestep to encourage the agent to complete tasks quickly. At the end of the episode, the agent is updated by looking over the past state representation when action choices were made and the resulting discounted reward for the scheduling decision made. These updates follow the traditional actor-critic losses and happen on-policy as the function approximator is not updated during an episode. Additionally, we use a decaying temperature in the SoftMax action selection to gradually reduce exploration and move between SoftMax to argmax. This is similar to the addition of entropy and lets the agent avoid skewed action distribution early in learning and introduces more exploration at the beginning of the simulation while slowly relying on exploitation later in training.
\begin{equation}\label{eq:pg}
    \nabla_\theta J(\theta) = \mathbb{E}_{\pi_{\theta}}[\nabla_{\theta}\log\pi_{\theta}(a_t|s_t)A_t(s_t)],
\end{equation}
Above, the objective is to find $\theta$ that parameterizes the neural network to maximize $J$. $A_t$ is the advantage that subtracts the state-value of state $s_t$, $V(s_t)$, from $G_t$, the empirically observed discounted reward, the baseline $V(s_t)$ serving to reduce potentially high variance. The above specifies the actor-loss in the actor-critic framework. The critic is updated to minimize the advantage, i.e., $(G_t - V(s_t))^2$ is minimized. The overall algorithm with DS3 is described in Algorithm~\ref{alg:drm}.

\begin{algorithm}
\begin{algorithmic}
   \STATE {\bfseries Input:} \texttt{jobs}, \texttt{resource\_matrices}, and DS3 environment
   \FOR{each episode}
    \STATE initialize environment with next \texttt{job} and \texttt{resource\_matrix} file
    \REPEAT 
    \FOR{tasks in ready list}
      \STATE Construct $state$
        \STATE Choose action $action$ w.r.t. task
        \STATE Assign $action$ to \texttt{PE\_ID} for this task
        \STATE Save $state$, $action$
    \ENDFOR
    \STATE Penalize -1 for $reward$
    \UNTIL{all tasks are the in the \texttt{completed\_list} or \texttt{max\_simulation\_length}}
    \STATE Compute losses using Eq.~\ref{eq:pg} with saved $state$s and $action$s
    \STATE Update agent by backpropagating with the losses
  \ENDFOR
\end{algorithmic}
 \caption{Deep Resource Management}
 \label{alg:drm}
\end{algorithm}

\section{Experiments}
\label{sec:experiments}
To the best of our knowledge, this paper is the first to apply reinforcement learning to heterogeneous resource management where long-term scheduling decisions need to be made. In this section, we show the experimental results of the comparison of the DRM scheduler with other heuristic schedulers, Earliest Finish Time (EFT), Earliest Time First (ETF), and Minimum Execution Time (MET)~\cite{buttazzo2011hard}. Both EFT and ETF pick the resource which is expected to give the earliest finish time for the task. EFT first come first served, whereas ETF looks over all available tasks and calculates their finishing time. MET assigns the processing element ID with minimum execution time for the current task.

\begin{figure}[ht]
    \centering
    \includegraphics[width=\linewidth]{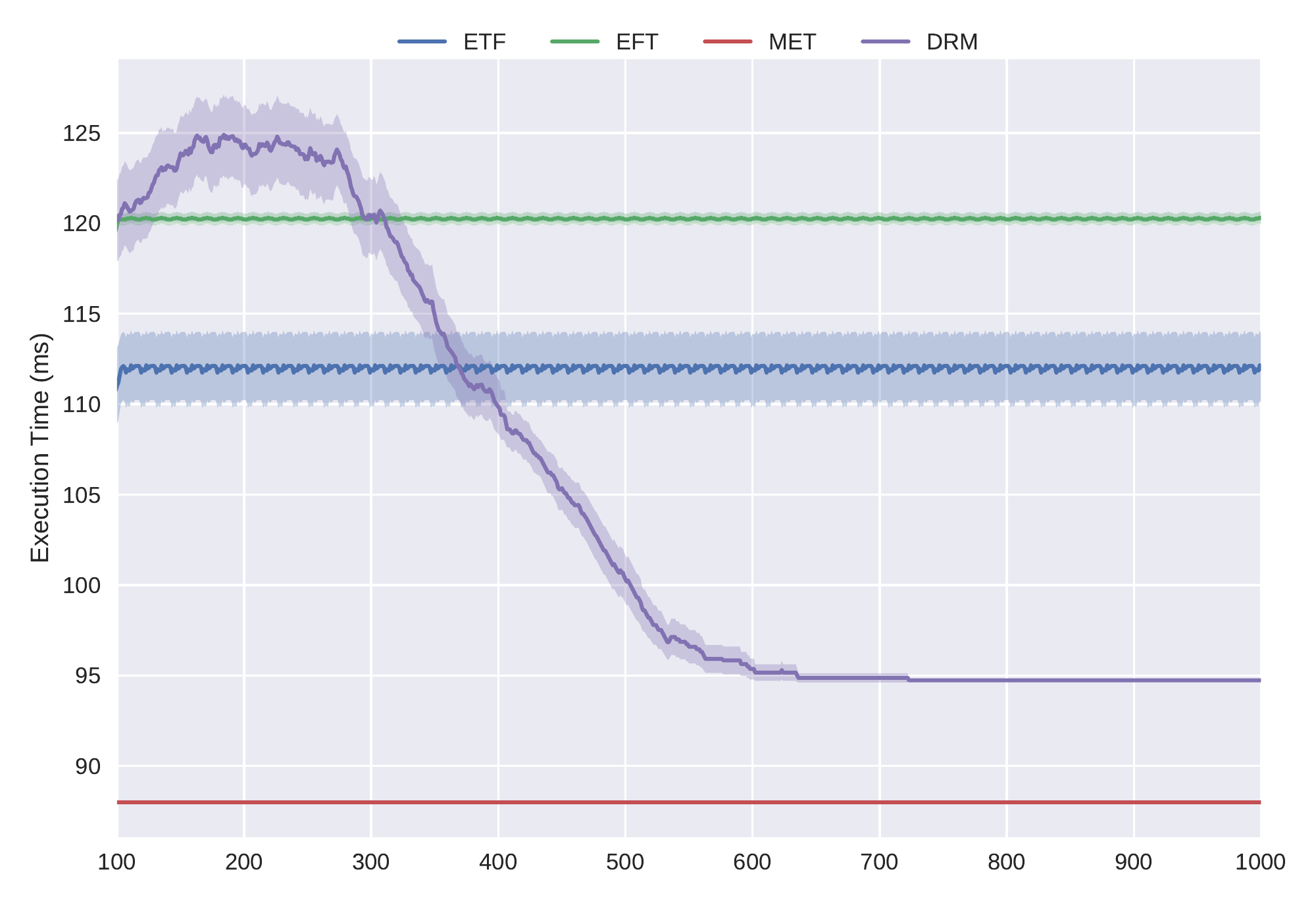}
    \centering
    \includegraphics[width=\linewidth]{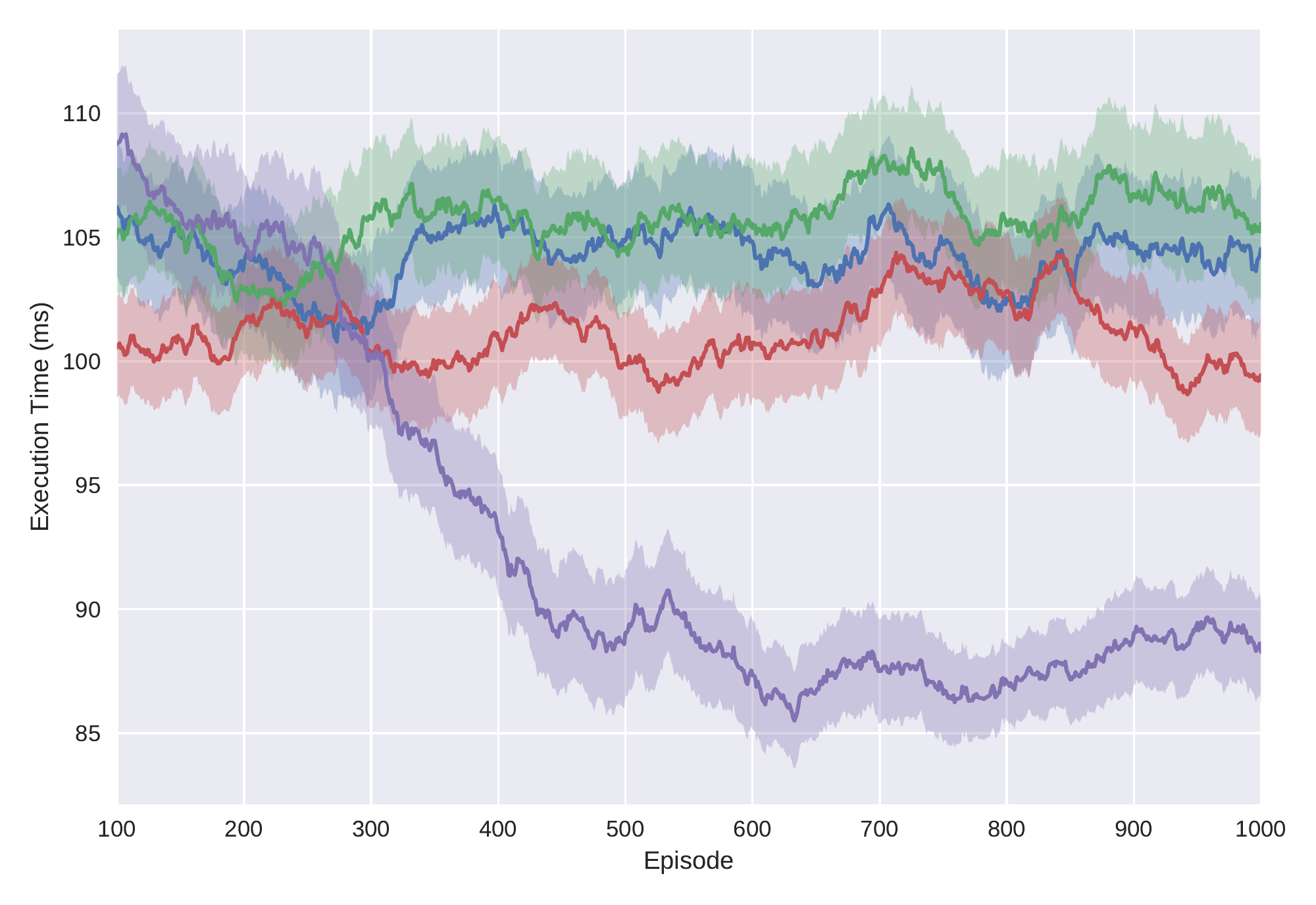}
    \caption{Execution Time versus Episode during training for the simpler case where the resource matrix specifying performance and communication characteristics is fixed (\textbf{Top}) and is randomized before each run (\textbf{Bottom})}
    \label{fig:RRM_ET_Plot}
\end{figure}

As shown in Figure~\ref{fig:RRM_ET_Plot}, the deep RL scheduler has the best performance overall. We tried different experiment settings: one with fixed data shown in top and another with randomized data in bottom. According to the fixed input, the DRM agent is trained to have about 94ms performance and saturated starting at about 720 episodes. Because the static data does not have much variation, the agent does not have much variance in performance but eventually overfits to a certain extent. Interestingly, MET has better performance than DRM agent, because MET picks a resource which has the minimum execution time for the tasks in ready list. We presume the MET corresponds to a locally optimal action at every timestep whereas DRM could not exceed this optimal value.

When we experimented with the randomized data, our DRM scheduler had the best performance. Despite fluctuating results of all schedulers, the DRM agent is the only one that had improved performance, of course, over time as learning progressed. The DRM agent applies an RL algorithm to explore various policies given different jobs and PEs, allowing for better generalization and better adaptivity. To provide  convincing results, we performed 30 trials with different random seeds. We expect to apply ensembles in the training experiences to provide much more reliable model. 

\begin{figure}[ht]
\begin{center}
\centerline{\includegraphics[width=\linewidth]{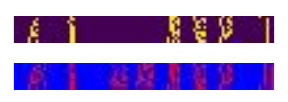}}
\caption{A visualization of the state representation which has information about task lists and PEs. Initialized state representation (\textbf{Top}) and the result of GradCam performed with the DRM feature layers (\textbf{Bottom}).}
\label{fig:gradcam}
\end{center}
\vskip -0.2in
\end{figure}

\begin{figure}[ht]
\begin{center}
\includegraphics[width=\linewidth, trim={3cm 0 3cm 0}, clip]{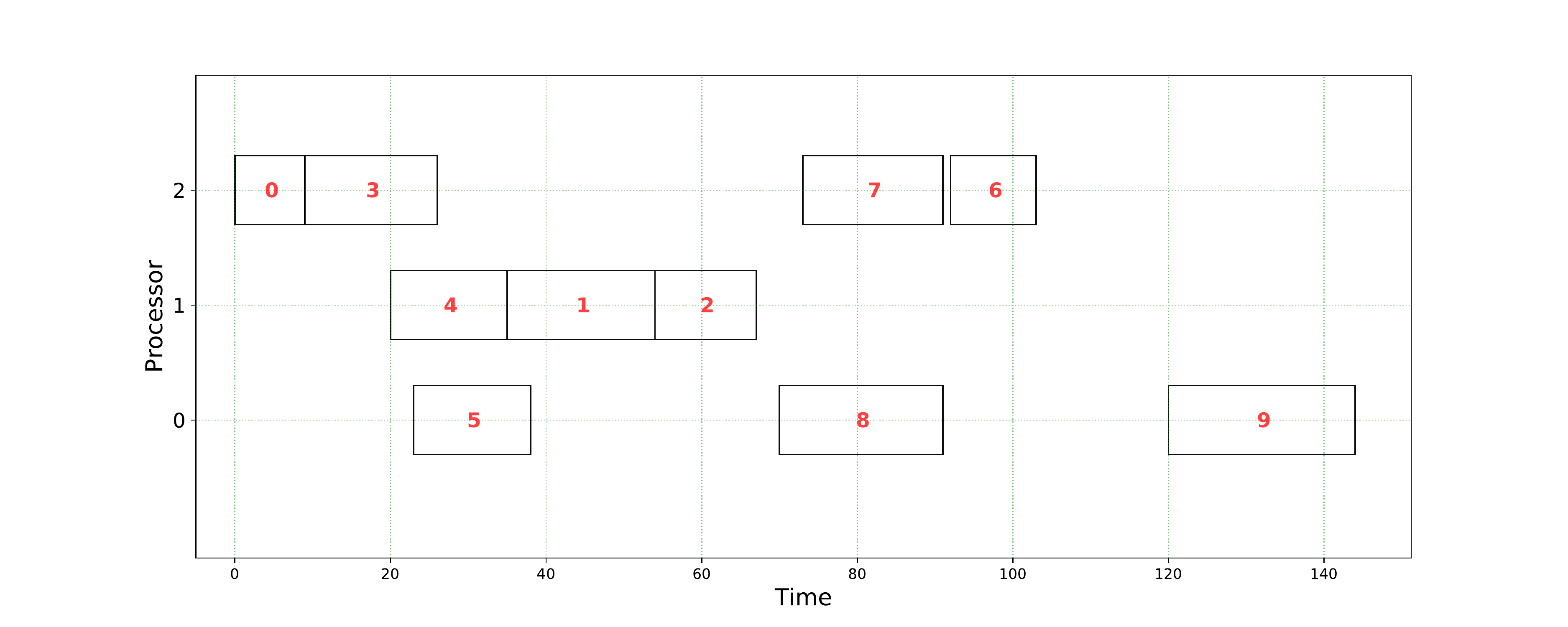}
\includegraphics[width=\linewidth, trim={3cm 0 3cm 0}, clip]{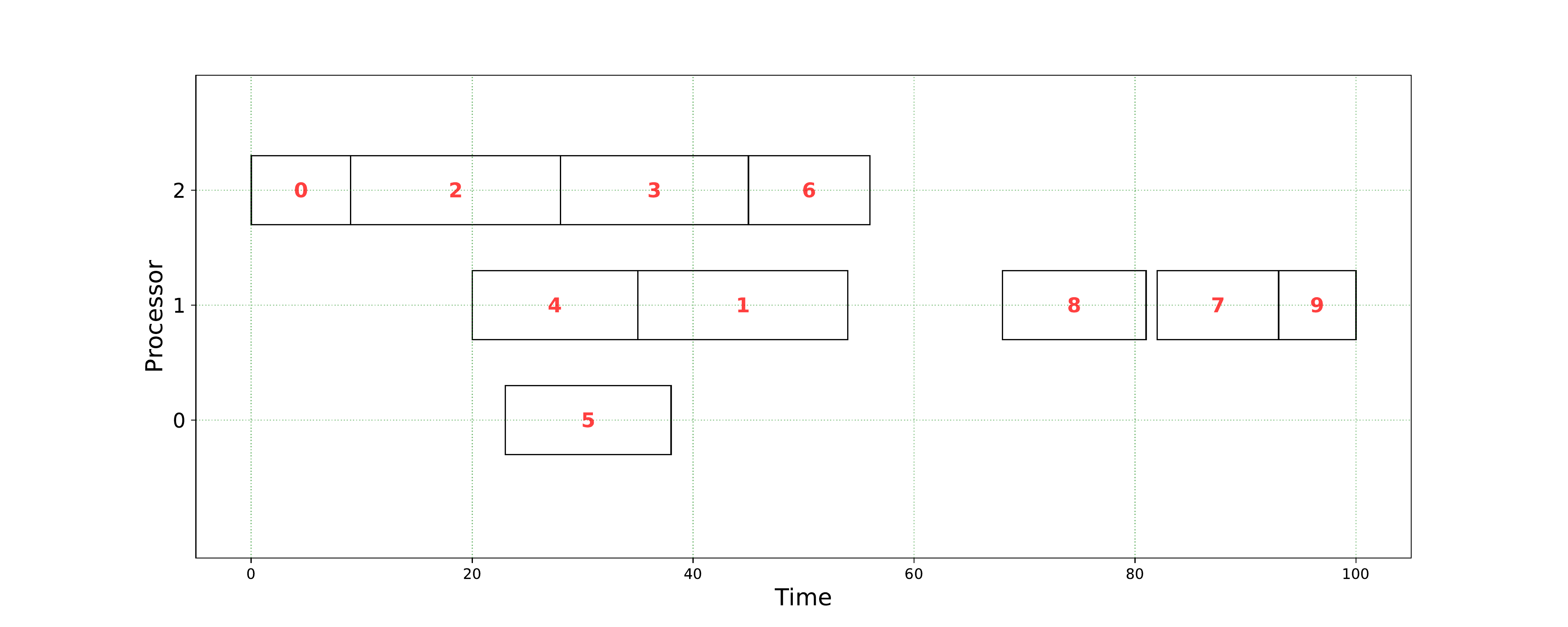}
\caption{GANTT chart representing when tasks ran on the different processing elements for the first episode (Top) and the last episode (Bottom) when training DRM}
\label{fig:gantt}
\end{center}
\vskip -0.2in
\end{figure}

We provide a visualization of the saliency to reason about the action decisions, shown in Figure~\ref{fig:gradcam}. The top figure shows the initial state representation formed by the task lists and resource matrix information. After passing the state to DRM agent, we perform GradCam~\cite{selvaraju2017grad} and retrieve the saliency, mapped onto the input, shown in the bottom of Figure~\ref{fig:gradcam}. Notice that the agent oversees tasks which are not shown in initial representation. Described with different color intensities, we presume that the DRM agent actually understand the tasks belonging to different status lists and that this more complex decision making input allows for better policies.

Finally, the GANTT chart showcases how the policy improves over training for the fixed resource matrix case. Initially, DRM gets quite a high execution time of 140 ms while it produces a better policy of about 100 ms at the end of training. Note that this chart corresponds to the task dependency graph shown earlier, Figure~\ref{fig:task-depend}, with the only difference being that the tasks are 0-indexed.

Some interesting changes include the choice of processing element 1 for Task 9 over processing element 0. It is cleat that this task is faster on PE1 compared to PE0. On the other hand, the choice of PE2 for task 2 vs PE1 is also interesting as task 2 takes longer. However, this might be better due to the task-dependency graph.

\section{Related Work}
\label{sec:related-work}
Resource management has been actively researched in many communities. Several works have been applying deep RL to optimally allocate resources, distribute tasks, and optimize power management decisions~\cite{gupta2019deep}. DeepRM uses standard deep Q-learning algorithm to formalize resource management as a Tetris game, however, it only work with homogeneous settings and not consider task dependency~\cite{mao2016resource}. A variant of DeepRM leverages convolution neural networks as a backbone network to improve performance in scheduling~\cite{chen2017deep}. Subsequent work in DeepRM, Pensieve applies a resource managing algorithm to video streaming to optimally control the bitrate and successfully reduced buffering~\cite{mao2017neural}. Moreover, Hopfield neural network has been applied to design heterogeneous multiprocessor architecture scheduler~\cite{chillet2011real}. More recent work combines heuristic and learning algorithms, starting from an existing plan and iteratively improving it and successfully applying it in heterogeneous job scheduling task~\cite{chen2018learning}. However, their work follows the general MDP setting where, again, the agent chooses action at every timestep. From the perspective of hardware, recent work has proposed new accelerator architectures which have potential advances~\cite{chen2018eyeriss}.

\section{Conclusion}
\label{sec:conclusion}
Neural schedulers using deep reinforcement learning have been researched in many areas and greatly improved the performance compared to heuristic algorithms. In this paper, we propose an approach of resource allocation applied in heterogeneous SoC chips. We use `no-operation' action and refer to all task lists, regardless of task status, to address partially-observability and the SMDP problem. To the best of our knowledge, this paper is the first to deal with scheduling different tasks on different hardware chips to discover the optimal combination of functionalities. We expect the general value functions and predictive knowledge approach and the option framework to improve performance and leave them as future work.

\section*{Acknowledgements}
\label{sec:acknowledgements}
This material is based on research sponsored by Air Force Research Laboratory (AFRL) and Defense Advanced Research Projects Agency (DARPA) under agreement number FA8650-18-2-7860. The U.S. Government is authorized to reproduce and distribute reprints for Governmental purposes notwithstanding any copyright notation thereon. The views and conclusions contained herein are those of the authors and should not be interpreted as necessarily representing the official policies or endorsements, either expressed or implied, of Air Force Research Laboratory (AFRL) and Defense Advanced Research Projects Agency (DARPA) or the U.S. Government. The authors would like to thank Samet Arda, Anish Krishnakumar, Umit Ogras and Daniel Bliss of Arizona State University for their constructive feedback and support for the research.


\begin{thebibliography}{22}
\providecommand{\natexlab}[1]{#1}
\providecommand{\url}[1]{\texttt{#1}}
\expandafter\ifx\csname urlstyle\endcsname\relax
  \providecommand{\doi}[1]{doi: #1}\else
  \providecommand{\doi}{doi: \begingroup \urlstyle{rm}\Url}\fi

\bibitem[Arda et~al.(2019)]{arda2019ds3}
Arda, S. et~al.
\newblock Ds3: A system-level domain-specific system-on-chip simulation
  framework.
\newblock Technical report, Arizona State University, June 2019.

\bibitem[Brockman et~al.(2016)Brockman, Cheung, Pettersson, Schneider,
  Schulman, Tang, and Zaremba]{brockman2016openai}
Brockman, G., Cheung, V., Pettersson, L., Schneider, J., Schulman, J., Tang,
  J., and Zaremba, W.
\newblock Openai gym.
\newblock \emph{arXiv preprint arXiv:1606.01540}, 2016.

\bibitem[Buttazzo(2011)]{buttazzo2011hard}
Buttazzo, G.~C.
\newblock \emph{Hard real-time computing systems: predictable scheduling
  algorithms and applications}, volume~24.
\newblock Springer Science \& Business Media, 2011.

\bibitem[Chen et~al.(2017)Chen, Xu, and Wu]{chen2017deep}
Chen, W., Xu, Y., and Wu, X.
\newblock Deep reinforcement learning for multi-resource multi-machine job
  scheduling.
\newblock \emph{arXiv preprint arXiv:1711.07440}, 2017.

\bibitem[Chen \& Tian(2018)Chen and Tian]{chen2018learning}
Chen, X. and Tian, Y.
\newblock Learning to progressively plan.
\newblock \emph{arXiv preprint arXiv:1810.00337}, 2018.

\bibitem[Chen et~al.(2018)Chen, Emer, and Sze]{chen2018eyeriss}
Chen, Y.-H., Emer, J., and Sze, V.
\newblock Eyeriss v2: A flexible and high-performance accelerator for emerging
  deep neural networks.
\newblock \emph{arXiv preprint arXiv:1807.07928}, 2018.

\bibitem[Chillet et~al.(2011)Chillet, Eiche, Pillement, and
  Sentieys]{chillet2011real}
Chillet, D., Eiche, A., Pillement, S., and Sentieys, O.
\newblock Real-time scheduling on heterogeneous system-on-chip architectures
  using an optimised artificial neural network.
\newblock \emph{Journal of Systems Architecture}, 57\penalty0 (4):\penalty0
  340--353, 2011.

\bibitem[Deng et~al.(2009)Deng, Dong, Socher, Li, Li, and
  Fei-Fei]{deng2009imagenet}
Deng, J., Dong, W., Socher, R., Li, L.-J., Li, K., and Fei-Fei, L.
\newblock Imagenet: A large-scale hierarchical image database.
\newblock In \emph{2009 IEEE conference on computer vision and pattern
  recognition}, pp.\  248--255. Ieee, 2009.

\bibitem[Gupta et~al.(2017)Gupta, Patil, Bhat, Mishra, and
  Ogras]{gupta2017dypo}
Gupta, U., Patil, C.~A., Bhat, G., Mishra, P., and Ogras, U.~Y.
\newblock Dypo: Dynamic pareto-optimal configuration selection for
  heterogeneous mpsocs.
\newblock \emph{ACM Transactions on Embedded Computing Systems (TECS)},
  16\penalty0 (5s):\penalty0 123, 2017.

\bibitem[Gupta et~al.(2019)Gupta, Mandal, Mao, Chakrabarti, and
  Ogras]{gupta2019deep}
Gupta, U., Mandal, S.~K., Mao, M., Chakrabarti, C., and Ogras, U.~Y.
\newblock A deep q-learning approach for dynamic management of heterogeneous
  processors.
\newblock \emph{IEEE Computer Architecture Letters}, 18\penalty0 (1):\penalty0
  14--17, 2019.

\bibitem[Krizhevsky et~al.(2012)Krizhevsky, Sutskever, and
  Hinton]{krizhevsky2012imagenet}
Krizhevsky, A., Sutskever, I., and Hinton, G.~E.
\newblock Imagenet classification with deep convolutional neural networks.
\newblock In \emph{Advances in neural information processing systems}, pp.\
  1097--1105, 2012.

\bibitem[Levine et~al.(2016)Levine, Finn, Darrell, and Abbeel]{levine2016end}
Levine, S., Finn, C., Darrell, T., and Abbeel, P.
\newblock End-to-end training of deep visuomotor policies.
\newblock \emph{The Journal of Machine Learning Research}, 17\penalty0
  (1):\penalty0 1334--1373, 2016.

\bibitem[Lünsdorf \& Scherfke(2018)Lünsdorf and Scherfke]{Simpy2018}
Lünsdorf, O. and Scherfke, S.
\newblock Simpy3.
\newblock \url{https://github.com/cristiklein/simpy}, 2018.

\bibitem[Mao et~al.(2016)Mao, Alizadeh, Menache, and Kandula]{mao2016resource}
Mao, H., Alizadeh, M., Menache, I., and Kandula, S.
\newblock Resource management with deep reinforcement learning.
\newblock In \emph{Proceedings of the 15th ACM Workshop on Hot Topics in
  Networks}, pp.\  50--56. ACM, 2016.

\bibitem[Mao et~al.(2017)Mao, Netravali, and Alizadeh]{mao2017neural}
Mao, H., Netravali, R., and Alizadeh, M.
\newblock Neural adaptive video streaming with pensieve.
\newblock In \emph{Proceedings of the Conference of the ACM Special Interest
  Group on Data Communication}, pp.\  197--210. ACM, 2017.

\bibitem[Puterman(2014)]{puterman2014markov}
Puterman, M.~L.
\newblock \emph{Markov decision processes: discrete stochastic dynamic
  programming}.
\newblock John Wiley \& Sons, 2014.

\bibitem[Selvaraju et~al.(2017)Selvaraju, Cogswell, Das, Vedantam, Parikh, and
  Batra]{selvaraju2017grad}
Selvaraju, R.~R., Cogswell, M., Das, A., Vedantam, R., Parikh, D., and Batra,
  D.
\newblock Grad-cam: Visual explanations from deep networks via gradient-based
  localization.
\newblock In \emph{Proceedings of the IEEE International Conference on Computer
  Vision}, pp.\  618--626, 2017.

\bibitem[Silver et~al.(2017)Silver, Schrittwieser, Simonyan, Antonoglou, Huang,
  Guez, Hubert, Baker, Lai, Bolton, et~al.]{silver2017mastering}
Silver, D., Schrittwieser, J., Simonyan, K., Antonoglou, I., Huang, A., Guez,
  A., Hubert, T., Baker, L., Lai, M., Bolton, A., et~al.
\newblock Mastering the game of go without human knowledge.
\newblock \emph{Nature}, 550\penalty0 (7676):\penalty0 354, 2017.

\bibitem[Sutton \& Barto(2018)Sutton and Barto]{sutton2018reinforcement}
Sutton, R.~S. and Barto, A.~G.
\newblock \emph{Reinforcement learning: An introduction}.
\newblock MIT press, 2018.

\bibitem[Sutton et~al.(2000)Sutton, McAllester, Singh, and
  Mansour]{sutton2000policy}
Sutton, R.~S., McAllester, D.~A., Singh, S.~P., and Mansour, Y.
\newblock Policy gradient methods for reinforcement learning with function
  approximation.
\newblock In \emph{Advances in neural information processing systems}, pp.\
  1057--1063, 2000.

\bibitem[Uhrie et~al.(2019)Uhrie, Bliss, Chakrabarti, Ogras, and
  Brunhaver]{uhrie2019machine}
Uhrie, R., Bliss, D.~W., Chakrabarti, C., Ogras, U.~Y., and Brunhaver, J.
\newblock Machine understanding of domain computation for domain-specific
  system-on-chips (dssoc).
\newblock In \emph{Open Architecture/Open Business Model Net-Centric Systems
  and Defense Transformation 2018}, volume 11015, pp.\  110150O. International
  Society for Optics and Photonics, 2019.

\bibitem[Vinyals et~al.(2019)Vinyals, Babuschkin, Chung, Mathieu, Jaderberg,
  Czarnecki, Dudzik, Huang, Georgiev, Powell, et~al.]{vinyals2019alphastar}
Vinyals, O., Babuschkin, I., Chung, J., Mathieu, M., Jaderberg, M., Czarnecki,
  W., Dudzik, A., Huang, A., Georgiev, P., Powell, R., et~al.
\newblock Alphastar: Mastering the real-time strategy game starcraft ii, 2019.

\end{thebibliography}

\bibliographystyle{icml2019}

\end{document}